\documentclass{article}
\usepackage{ijcai15}

\usepackage{times}
\usepackage[fleqn]{amsmath}
\usepackage{epsf}
\usepackage{graphicx}
\usepackage{epstopdf}

\title{Medical Synonym Extraction with Concept Space Models}

\author{Chang Wang \and  Liangliang Cao \and  Bowen Zhou\\
IBM T. J. Watson Research Lab\\
1101 Kitchawan Rd\\
Yorktown Heights, New York 10598\\
\{\texttt{changwangnk},
\texttt{liangliang.cao}\}\texttt{@gmail.com},
\texttt{zhou}\texttt{@us.ibm.com}}

\begin{document}

\maketitle

\begin{abstract}
In this paper, we present a novel approach for medical synonym
extraction. We aim to integrate the term embedding with the
medical domain knowledge for healthcare applications. One
advantage of our method is that it is very scalable. Experiments
on a dataset with more than 1M term pairs show that the proposed
approach outperforms the baseline approaches by a large margin.

\end{abstract}

\section{Introduction}

Many components to build a high quality natural language
processing system rely on the synonym extraction. Examples include
query expansion, text summarization~\cite{textsummarization},
question answering~\cite{thisiswatson-12}, and paraphrase
detection. Although the value of synonym extraction is undisputed,
manual construction of such resources is always expensive, leading
to a low knowledgebase (KB) coverage~\cite{medicalsynonym}.

In the medical domain, this KB coverage issue is more serious,
since the language use variability is exceptionally
high~\cite{medicaliereview}. In addition, the natural language
content in the medical domain is also growing at an extremely high
speed, making people hard to understand it, and update it in the
knowledgebase in a timely manner.

To construct a large scale medical synonym extraction system, the
main challenge to address is how to build a system that can
automatically combine the existing manually extracted medical
knowledge with the huge amount of the knowledge buried in the
unstructured text. In this paper, we construct a medical corpus
containing 130M sentences (20 gigabytes pure text). We also
construct a semi-supervised framework to generate a vector
representation for each medical term in this corpus. Our framework
extends the Word2Vec model~\cite{word2vec} by integrating the
existing medical knowledge in the model training process.

To model the concept of synonym, we build a ``\textbf{concept
space}'' that contains both the semi-supervised term embedding
features and the expanded features that capture the similarity of
two terms on both the word embedding space and the surface form.
We then apply a linear classifier directly to this space for
synonym extraction. Since both the manually extracted medical
knowledge and the knowledge buried under the unstructured text
have been encoded in the concept space, a cheap classifier can
produce satisfying extraction results, making it possible to
efficiently process a huge amount of the term pairs.

Our system is designed in such a way that both the existing
medical knowledge and the context in the unstructured text are
used in the training process. The system can be directly applied
to the input term pairs without considering the context. The
overall contributions of this paper on medical synonym extraction
are two-fold:

\begin{itemize}
\item {From the perspective of applications, we identify a number
of Unified Medical Language System (UMLS)~\cite{umls} relations
that can be mapped to the synonym relation (Table
\ref{tbl:synonymrelations}), and present an automatic approach to
collect a large amount of the training and test data for this
application. We also apply our model to a set of 11B medical term
pairs, resulting in a new medical synonym knowledgebase with more
than 3M synonym candidates unseen in the previous medical
resources.}

\item {From the perspective of methodologies, we present a
semi-supervised term embedding approach that can train the vector
space model using both the existing medical domain knowledge and
the text data in a large corpus. We also expand the term embedding
features to form a concept space, and use it to facilitate synonym
extraction.}

\end{itemize}

The experimental results show that our synonym extraction models
are fast and outperform the state-of-the-art approaches on medical
synonym extraction by a large margin. The resulting synonym KB can
also be used as a complement to the existing knowledgebases in
information extraction tasks.

\section{Related Work}\label{sec:relatedwork}

A wide range of techniques has been applied to synonym detection,
including the use of lexicosyntactic patterns~\cite{synonym-11},
clustering~\cite{brown-cluster}, graph-based
models~\cite{synonym-12}\cite{synonym-13}\cite{synonym-tong}\cite{synonym-minkov}
and distributional
semantics~\cite{synonym-10}\cite{medicalsynonym}\cite{synonym-24}\cite{synonym-25}\cite{synonym-28}.
There are also efforts to improve the detection performance using
multiple sources or ensemble
methods~\cite{synonym-15}\cite{synonym-16}\cite{synonym-18}.

The vector space models are directly related to synonym
extraction. Some approaches use the low rank approximation idea to
decompose large matrices that capture the statistical information
of the corpus. The most representative method under this category
is Latent Semantic Analysis (LSA)~\cite{lsi}. Some new models also
follow this approach like Hellinger PCA~\cite{hpca} and
GloVe~\cite{global-word2vec}.

Neural network based representation learning has attracted a lot
of attentions recently. One of the earliest work was done in
\cite{Rumelhart-nn}. This idea was then applied to language
modeling~\cite{bengio-nnlanguagemodel}, which motivated a number
of research projects in machine learning to construct the vector
representations for natural language processing tasks~
\cite{collobert-nn}\cite{turney-nn}\cite{glorot-nn}\cite{socher-rnn}\cite{mnih-nn}\cite{andrew-word2vec}.

Following the same neural network language modeling idea,
Word2Vec~\cite{word2vec} significantly simplifies the previous
models, and becomes one of the most efficient approach to learn
word embeddings. In Word2Vec, there are two ways to generate the
``input-desired output'' pairs from the context: ``SkipGram''
(predicts the surrounding words given the current word) and
``CBOW'' (predicts the current word based on the surrounding
words), and two approaches to simplify the training: ``Negative
Sampling'' (the vocabulary is represented in one hot
representation, and the algorithm only takes a number of randomly
sampled ``negative'' examples into consideration at each training
step) and ``Hierarchical SoftMax'' (the vocabulary is represented
as a Huffman binary tree). So one can train a Word2Vec model from
the input under 4 different settings, like ``SkipGram''+``Negative
Sampling". Word2Vec is the basis of our semi-supervised word
embedding model, and we will discuss it with more details in
Section~\ref{sec:conceptembedding}.

\section{Medical Corpus, Concepts and Synonyms}

\subsection{Medical Corpus}\label{sec:corpus}

Our medical corpus has incorporated a set of Wikipedia articles
and MEDLINE abstracts (2013
version)\footnote{http://www.nlm.nih.gov/bsd/pmresources.html}. We
also complemented these sources with around 20 medical journals
and books like \emph{Merck Manual of Diagnosis and Therapy}. In
total, the corpus contains about 130M sentences (about 20G pure
text), and about 15M distinct terms in the vocabulary set.

\subsection{Medical Concepts}\label{sec:concepts} A significant amount of the medical
knowledge has already been stored in the Unified Medical Language
System (UMLS)~\cite{umls}, which includes medical concepts,
definitions, relations, etc. The 2012 version of the UMLS contains
more than 2.7 million concepts from over 160 source vocabularies.
Each concept is associated with a unique keyword called CUI
(Concept Unique Identifier), and each CUI is associated with a
term called preferred name. The UMLS consists of a set of 133
subject categories, or semantic types, that provide a consistent
categorization of all CUIs. The semantic types can be further
grouped into 15 semantic groups. These semantic groups provide a
partition of the UMLS Metathesaurus for 99.5\% of the concepts.

Domain specific parsers are required to accurately process the
medical text. The most well-known parsers in this area include
MetaMap (Aronson, 2001) and MedicalESG, an adaptation of the
English Slot Grammar parser~\cite{esg-12} to the medical domain.
These tools can detect medical entity mentions in a given
sentence, and automatically associate each term with a number of
CUIs. Not all the CUIs are actively used in the medical text. For
example, only 150K CUIs have been identified by MedicalESG in our
corpus, even though there are in total 2.7M CUIs in UMLS.

\subsection{Medical Synonyms}\label{sec:medicalsynonyms}

Synonymy is a semantic relation between two terms with very
similar meaning. However, it is extremely rare that two terms have
the exact same meaning. In this paper, our focus is to identify
the near-synonyms, i.e. two terms are interchangeable in some
contexts~\cite{semantics}.

The UMLS 2012 Release contains more than 600 relations and 50M
relation instances under 15 categories. Each category covers a
number of relations, and each relation has a certain number of CUI
pairs that are known to bear that relation. From UMLS relations,
we manually choose a subset of them that are directly related to
synonyms, and summarize them in Table~\ref{tbl:synonymrelations}.
In Table~\ref{tbl:synonymexamples}, we list several synonym
examples provided by these relations.

\begin{table}[!ht] \centering\caption{UMLS Relations Corresponding
to the Synonym Relation, where ``RO'' stands for ``has
relationship other than synonymous, narrower, or broader'', ``RQ''
stands for ``related and possibly synonymous'', and ``SY'' stands
for ``source asserted synonymy''.} \label{tbl:synonymrelations}
\begin{tabular}{|c|c|}
\hline Category & Relation Attribute\\
\hline RO & has\_active\_ingredient \\
\hline RO & has\_ingredient \\
\hline RO & has\_product\_component \\
\hline RO & has\_tradename \\
\hline RO & refers\_to \\
\hline RQ & has\_alias \\
\hline RQ & replaces \\
\hline SY & -\\
\hline SY & expanded\_form\_of\\
\hline SY & has\_expanded\_form \\
\hline SY & has\_multi\_level\_category \\
\hline SY & has\_print\_name \\
\hline SY & has\_single\_level\_category\\
\hline SY & has\_tradename \\
\hline SY & same\_as \\
\hline
\end{tabular}
\end{table}

\begin{table}[!ht] \centering\caption{Several Synonym Examples in UMLS.} \label{tbl:synonymexamples}
\begin{tabular}{|c|c|}
\hline term 1 & term 2 \\
\hline Certolizumab pegol & Cimzia  \\
\hline Child sex abuse (finding) & Child Sexual Abuse \\
\hline Mass of body structure & Space-occupying mass \\
\hline Multivitamin tablet & Vitamin A \\
\hline Blood in stool & Hematochezia \\
\hline Keep Alert & Caffeine \\
\hline
\end{tabular}
\end{table}

\section{Medical Synonym Extraction with Concept Space Models}
In this section, we first project the terms to a new vector space,
resulting in a vector representation for each term
(Section~\ref{sec:conceptembedding}). This is done by adding the
semantic type and semantic group knowledge from UMLS as extra
labels to the Word2Vec model training process. In the second step
(Section~\ref{sec:featureexpansion}), we expand the resulted term
vectors with extra features to model the relationship between two
terms. All these features together form a \emph{\textbf{concept
space}} and are used with a linear classifier for medical synonym
extraction.

\subsection{Semi-Supervised Term
Embedding}\label{sec:conceptembedding}

\subsubsection{High Level Idea}

A key point of the Word2Vec model is to generate a ``desired
word'' for each term or term sequence from the context, and then
learn a shallow neural network to predict this ``term
$\longrightarrow$ desired word'' or ``term sequence
$\longrightarrow$ desired word'' mapping. The ``desired word''
generation process does not require any supervised information, so
it can be applied to any domain.

\begin{figure}
\includegraphics[width=0.375\textwidth]{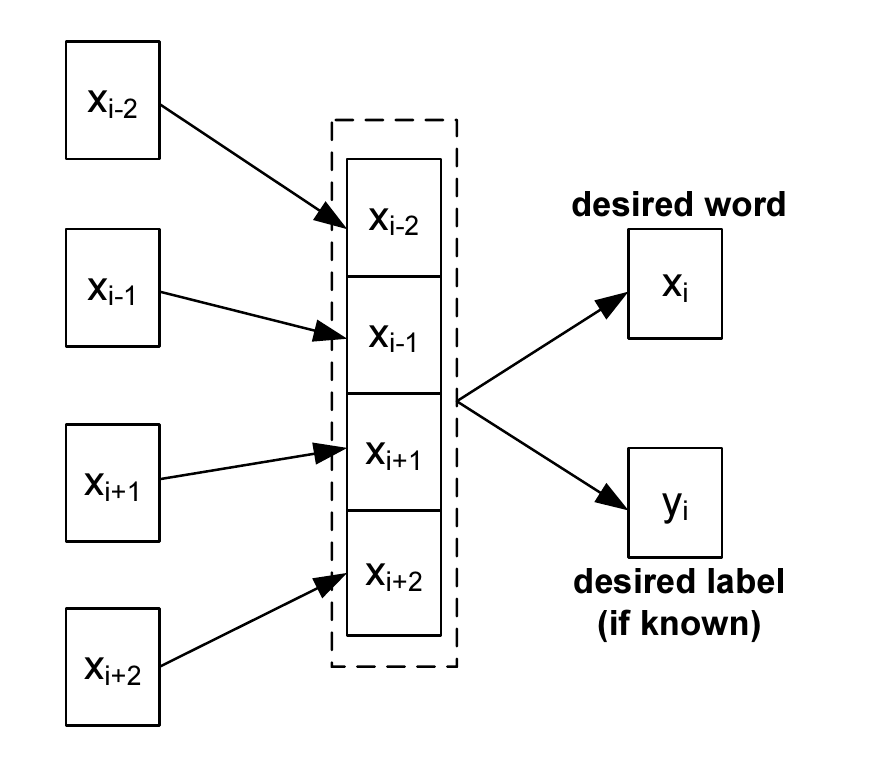}
\centering \caption{An Illustration of the Semi-supervised Word
Embedding Model.}\label{fig:illlustration}
\end{figure}

In the medical domain, a huge amount of the true label information
in the format of ``semantic type'', ``semantic group'', etc has
already be manually extracted and integrated in the knowledgebases
like UMLS. Our Semi-supervised approach is to integrate this type
of ``true label'' and the ``desired word'' in the neural network
training process to produce a better vector representation for
each term. This idea is illustrated in
Figure~\ref{fig:illlustration}.

\subsubsection{The Training Algorithm}

The training algorithm used in the Word2Vec model first learns how
to update the weights on the network edges assuming the word
vectors are given (randomly initialized); then it learns how to
update the word vectors assuming the weights on the edges are
given (randomly initialized). It repeats these two steps until all
the words in the input corpus have been processed.

We produce a ``true label'' vector, and add it as an extra set of
output nodes to the neural network. This vector has 148 entries
corresponding to 133 medical semantic types and 15 semantic
groups. If a term is associated with some semantic types and
semantic groups, the corresponding entries will be set to 1,
otherwise 0. Increasing the size of the output layer will slow
down the computation. For example, compared to the original
Word2Vec
 ``Negative Sampling'' strategy with 150 negative samples, the
semi-supervised method will be about $50\%$ slower.

This section derives new update rules for word vectors and edge
weights. Since the semi-supervised approach is an extension of the
original Word2vec model, the derivation is also similar. The new
update rules work for all 4 Word2Vec settings mentioned in
Section~\ref{sec:relatedwork}. The notations used in this section
are defined in Figure~\ref{fig:alg}.

\begin{figure}[!ht]
\center
\begin{tabular}{|p{7.8cm}|}
\hline


\item[] {$\delta(x)= 1/(1+e^{-x})$;}

\item[]

\item[] {$V$ is a vocabulary set with $|V|$ distinct words;}

\item[] $W= \{w_1 \cdots w_{|V|}\}$ is the set of embeddings for
all the words in $V$;

\item[] $w_k$ is the embedding for word $k$ in $V$;

\item[] $\theta_k$ is a mapping function associated with word $k$;

\item[]

\item[] $X= \{x_1, \cdots x_N\}$ represents all the words in a
corpus $\mathcal{X}$;

\item[] $x_i$ is the $i^{th}$ word in $X$;

\item[] $c_i= [w_{c_i^1}, \cdots w_{c_i^l}]$ represents the
embeddings of all $l$ context words for word $x_i$ in $X$;

\item[]

\item[] $L$ is a label set with $|L|$ distinct labels;

\item[] $X_U$ represents all the words without known labels in
$X$;

\item[] $X_L$ represents all the words with known labels in $X$;

\item[] $X= \{X_L, X_U\}$;

\item[] $Y_i \in L$ is a set of known labels associated with
$x_i$; Note that $x_i$ may have multiple labels;

\item[] $Y_i = \{\}$, when $x_i$'s label is unknown;

\item[]

\item[] $\eta$ is the learning rate;

\\\hline
\end{tabular}
\caption{Notations}\label{fig:alg}
\end{figure}
It can be verified that the following formulas will always hold:
\begin{eqnarray*}
\partial (\delta(x))/ \partial x = \delta(x) \cdot (1- \delta(x))
\end{eqnarray*}
\begin{eqnarray*}
\partial (\log \delta(x))/ \partial x = 1-\delta(x)
\end{eqnarray*}
\begin{eqnarray*}
\partial (\log (1-\delta(x)))/ \partial x = -\delta(x)
\end{eqnarray*}
\\$\pi_S(x)$ is defined as
\begin{eqnarray*}\pi_{S}(x)=
      \left\{
       \begin{array}{c}
        1, \,\,\,x \in S;\\
        0, \,\,\,other wise;\\
       \end{array}
      \right.
\end{eqnarray*}
\\
Given a word $x_i$, we first compute the context vector $c_i$ by
concatenating the embeddings of all the contextual words of $x_i$.
Then we simultaneously maximize the probability to predict its
``desired word'' and the ``true label'' (if known).
\begin{eqnarray*}f(u|c_i)=
      \left\{
       \begin{array}{c}
       \delta(\theta_u' c_i), \pi_{\{x_i, Y_i\}}(u)=1 \\
       1-\delta(\theta_u' c_i), \pi_{\{x_i, Y_i\}}(u)=0\\
       \end{array}
      \right.
\end{eqnarray*}
\\$f(u|c_i)$ can also be written as
\begin{eqnarray*}f(u|c_i)=
\delta(\theta_u' c_i)^{\pi_{\{x_i, Y_i\}}(u)}\cdot
(1-\delta(\theta_u' c_i))^{1-\pi_{\{x_i, Y_i\}}(u)}
\end{eqnarray*}
\\To make the explanation simpler, we define $F(x_i, u)$, $F(x_i)$ and $F$ as follows:
\begin{displaymath}
 F(x_i, u) = f(u|c_i),\,\,\,
 F(x_i) = \prod_{u \in V} f(u|c_i), \,\,\,
 F= \prod_{x_i \in X} F(x_i)
\end{displaymath}
\\
To achieve our goal, we want to maximize $F$ for the given $X$:
\\
\begin{eqnarray*}
F= \prod_{x_i \in X} \prod_{u \in V} f(u|c_i)
\end{eqnarray*}
\\Apply $\log$ to both sides of the above equation, we have
\begin{eqnarray*}
\log F&=& \sum_{x_i \in X} \sum_{u \in V} \log f(u|c_i) 
\\
&=&\sum_{x_i \in X} \sum_{u \in V} \log [\delta(\theta_u'
c_i)^{\pi_{\{x_i, Y_i\}}(u)} \\&\cdot&
(1-\delta(\theta_u'c_i))^{1-\pi_{\{x_i, Y_i\}}(u)}]\\
&=&\sum_{x_i \in X} \sum_{u \in V} \pi_{\{x_i, Y_i\}}(u) \cdot
\log \delta(\theta_u' c_i) \\&+&  (1-\pi_{\{x_i, Y_i\}}(u)) \cdot
\log (1-\delta(\theta_u'c_i))
\end{eqnarray*}
\\Now we compute the first derivative of $\log F(x_i, u)$ on both
$\theta_u$ and $c_i$:
\\$\partial \log F(x_i,u)/\partial \theta_u$
\begin{eqnarray*}
&=& \pi_{\{x_i, Y_i\}}(u) \cdot (1- \delta(\theta_u' c_i)) \cdot
c_i \\&-&  (1-\pi_{\{x_i, Y_i\}}(u)) \cdot
\delta(\theta_u'c_i) \cdot c_i\\
&=&  (\pi_{\{x_i, Y_i\}}(u) - \delta(\theta_u'c_i)) \cdot c_i
\end{eqnarray*}
\\$\partial \log F(x_i,u)/\partial c_i$
\begin{eqnarray*}
=(\pi_{\{x_i, Y_i\}}(u) - \delta(\theta_u'c_i)) \cdot \theta_u
\end{eqnarray*}
\\The new update rules to update $\theta_u$
(weights on the edges) and $c_i$ (word vectors) are as follows:
\begin{eqnarray}
\theta_u&=& \theta_u+\eta \cdot (\pi_{\{x_i, Y_i\}}(u) -
\delta(\theta_u'c_i)) \cdot c_i\\
c_i&=& c_i +\eta \cdot (\pi_{\{x_i, Y_i\}}(u) -
\delta(\theta_u'c_i)) \cdot \theta_u
\end{eqnarray}

\subsection{Expansion of the Raw Features}\label{sec:featureexpansion}

The semi-supervised term embedding model returns us with the
vector representation for each term. To capture more information
useful for synonym extraction, we expand the raw features with
several heuristic rule-based matching features and several other
simple feature expansions. All these expanded features together
with the raw features form a \textbf{concept space} for synonym
extraction.

Notations used in this section are summarized here: for any pair
of the input terms: $a$ and $b$, we represent their lengths
(number of words) as $|a|$ and $|b|$. $a$ and $b$ can be
multi-word terms (like ``United States'') or single-word terms
(like ``student''). We represent the raw feature vector of $a$ as
$A$, and the raw feature vector of $b$ as $B$.

\subsubsection{Rule-based Matching Features}
The raw features can only help model the distributional similarity
of two terms based on the corpus and the existing medical
knowledge. In this section, we provide several matching features
to model the similarity of two terms based on the surface form.
These rule-based matching features $m_1$ to $m_6$ are generated as
follows:

\begin{itemize}
\item $m_1$: returns the number of the common words shared by $a$
and $b$.

\item $m_2$: $m_1/(|a|\cdot|b|)$;

\item $m_3$: if $a$ and $b$ only differ by an antonym prefix,
returns 1; otherwise, 0. The antonym prefix list includes
character sequences like ``anti'', ``dis'', ``il'', ``im'',
``in'', ``ir'' ``non'' and ``un'', et al. For example, $m_3=1$ for
``like'' and ``dislike''.

\item $m_4$: if all the upper case characters from $a$ and $b$
match each other, returns 1; otherwise, 0. For example, $m_4=1$
for ``USA'' and "United States of America".

\item $m_5$: if all the first characters in each word from $a$ and
$b$ match each other, returns 1; otherwise, 0. For example,
$m_5=1$ for ``hs'' and ``hierarchical softmax''.

\item $m_6$: if one term is the subsequence of another term,
returns 1; otherwise, 0.

\end{itemize}

\subsubsection{Feature Expansions}
In addition to the matching features, we also produce several
other straightforward expansions, including

\begin{itemize}
\item ``sum'': $[A+B]$

\item ``difference'': $[\,|A-B|\,]$

\item ``product'': $[A \cdot B]$

\item $ [m_2 \cdot A \,\,,\,\, m_2 \cdot B]$

\end{itemize}

\section{Experiments}

In our experiments, we used the MedicalESG parser~\cite{esg-12} to
parse all 130M sentences in the corpus, and extracted all the
terms from the sentences. For those terms that are associated with
CUIs, we assigned each of them with a set of semantic types/groups
through a CUI lookup in UMLS.

Section~\ref{sec:datageneration} presents how the training and
test sets were created. Section~\ref{sec:baseline} is about the
baseline approaches used in our experiments. In
Section~\ref{sec:classficationresults}, we compare the new models
and the baseline approaches on medical synonym extraction task. To
measure the scalability, we also used our best synonym extractor
to build a new medical synonym knowledgebase. Then we analyze the
contribution of each individual feature to the final results in
Section~\ref{sec:contributionresults}.

\subsection{Data Collection}\label{sec:datageneration}
UMLS has about 300K CUI pairs under the relations (see
Table~\ref{tbl:synonymrelations}) corresponding to synonyms.
However, the majority of them contain the CUIs that are not in our
corpus. The goal of this paper is to detect new synonyms from
text, so only the relation instances with both CUIs in the corpus
are used in the experiments. A CUI starts with letter `C', and is
followed by 7 digits. We use the preferred name, which is a term,
as the surface form for each CUI. Our preprocessing step resulted
in a set of 8,000 positive synonym examples. To resemble the
real-world challenges, where most of the given term pairs are
non-synonyms, we randomly generated more than 1.6M term pairs as
the negative examples. For these negative examples, both terms are
required to occur at least twice in our corpus. Some negative
examples generated in this way may be in fact positives, but this
should be very rare.

The final dataset was split into 3 parts: 60\% examples were used
for training, 20\% were used for testing the classifiers, and the
remaining 20\% were held out to evaluate the knowledgebase
construction results.

\subsection{Baseline Approaches}\label{sec:baseline}

Both the LSA model~\cite{lsi} and the Word2Vec
model~\cite{word2vec} were built on our medical corpus as
discussed in Section~\ref{sec:corpus}, which has about 130M
sentences and 15M unique terms. We constructed Word2Vec as well as
the semi-supervised term embedding models under all 4 different
settings: HS+CBOW, HS+SkipGram, NEG+CBOW, and NEG+SkipGram. The
parameters used in the experiments were: dimension\_size=100,
window\_size=5, negative=10, and sample\_rate=1e-5. We obtained
100 dimensional embeddings from all these models. Word2Vec models
typically took a couple of hours to train, while LSA model
required a whole day training on a computer with 16 cores and 128G
memory. The feature expansion (Section~\ref{sec:featureexpansion})
was applied to all these baseline approaches as well.

The letter $n$-gram model~\cite{letterngram} used in our
experiments was slightly different from the original one. We added
special letter $n$-grams on top of the original one to model the
begin and end of each term. In our letter $n$-gram experiments, we
tested $n=2,3$ and $4$. The letter $n$-gram model does not require
training.

\subsection{Synonym Extraction Results}\label{sec:classficationresults}
The focus of this paper is to find a good concept space for
synonym extraction, so we prefer a simple classifier over a
complicated one in order to more directly measure the impact of
the features (in the concept space) on the performance. The speed
is one of our major concerns, and we have about 1M training
examples to process, so we used the liblinear
package~\cite{liblinear} in our experiments for its high speed and
good scalability. In all the experiments, the weight for the
positive examples was set to 100, due to the fact that most of the
input examples were negative. All the other parameters were set to
the default values. The evaluation of different approaches is
based on the $F_1$ scores, and the final results are summarized in
Table~\ref{tbl:comparisons}.
\begin{table}[!ht] \centering\caption{$F_1$ Scores of All Approaches on both the Training and the Test Data.} \label{tbl:comparisons}
\begin{tabular}{|c|c|c|}
\hline Approach                & {Training $F_1$}    & {Test $F_1$} \\
\hline LSA                     & 48.98\%  & 48.13\%\\
\hline Letter-BiGram           & 55.51\%  & 50.09\%\\
\hline Letter-TriGram          & 96.30\%  & 63.37\%\\
\hline Letter-FourGram         & \textbf{99.48\%}  & 66.50\%\\
\hline Word2Vec HS+CBOW    & 64.05\%  & 63.51\%\\
\hline Word2Vec HS+SKIP    & 64.23\%  & 62.65\%\\
\hline Word2Vec NEG+CBOW   & 59.47\%  & 58.75\%\\
\hline Word2Vec NEG+SKIP   & 70.17\%  & 67.86\%\\
\hline Concept Space HS+CBOW & 68.15\%  & 65.73\%\\
\hline Concept Space HS+SKIP & 74.25\%  & 70.74\%\\
\hline Concept Space NEG+CBOW& 63.57\%  & 60.90\%\\
\hline Concept Space NEG+SKIP& 74.09\%  & \textbf{70.97\%}\\
\hline
\end{tabular}
\end{table}

From the results in Table~\ref{tbl:comparisons}, we can see that
the LSA model returns the lowest $F_1$ score of $48.13\%$,
followed by the letter bigram model ($50.09\%$). The letter
trigram and 4-gram models return very competitive scores as high
as $66.50\%$, and it looks like increasing the value of $n$ in the
letter-$n$gram model will push the $F_1$ score up even further.
However, we have to stop at $n=4$ for two reasons. Firstly, when
$n$ is larger, the model is more likely to overfit for the
training data, and the $F_1$ score for the letter 4-gram model on
the training data is already $99.48\%$. Secondly, the resulting
model will be more complicated for $n>4$. For the letter 4-gram
model, the model file itself is already about 3G big on the disk,
making it very expensive to use.

The best setting of the Word2Vec model returns a $67.86\%$ $F_1$
score. This is about $3\%$ lower than the best setting of the
concept space model, which achieves the best $F_1$ score across
all approaches: $70.97\%$. We also compare the Word2Vec model and
the concept space model under all 4 different settings in
Table~\ref{tbl:unsemicomparisons}. The concept space model
outperforms the original Word2Vec model by a large margin ($3.9\%$
on average) under all of them.

Since we have a lot of training data, and are using a linear model
as the classifier, the training part is very stable. We ran a
couple of other experiments (not reported in this paper) by
expanding the training set and the test set with the dataset held
out for knowledgebase evaluation, but did not see too much
difference in terms of the $F_1$ scores.

\begin{table}[!ht] \centering\caption{$F_1$ Scores of the Word2Vec and the Concept Space Models on the Test Data.} \label{tbl:unsemicomparisons}
\begin{tabular}{|c|c|c|c|}
\hline Setting   & {Wored2Vec}    & {Concept Space} & {Diff}\\
\hline HS+CBOW   & 63.51\%            & 65.73\%           & +2.22\%\\
\hline HS+SKIP   & 62.65\%            & 70.74\%           & +8.09\%\\
\hline NEG+CBOW  & 58.75\%            & 60.90\%           & +2.15\%\\
\hline NEG+SKIP  & 67.86\%            & \textbf{70.97\%}  & +3.11\%\\
\hline Average   & 63.19\%            & 67.09\%           & +3.90\%\\
 \hline
\end{tabular}
\end{table}
\begin{table*}[!ht] \centering\caption{Analysis of the Feature
Contributions to $F_1$ for the Concept Space Model.}
\label{tbl:featurecontributions}
\begin{tabular}{|c|c|c|c|c|c|c|}
\hline \textbf{$F_1$}                     &{HS+}   &{HS+}   &{NEG+}  &{NEG+}           &{Average} &{Average}\\
                                          &{CBOW}  &{SKIP}  &{CBOW}  &{SKIP}           &{Score}   &{Improvement}\\
\hline $[A, B]$                           & 44.28\% & 51.19\% & 41.42\% & 53.37\%          & 47.57\%   & - \\
\hline + [matching features]              & 57.81\% & 63.53\% & 54.63\% & 62.37\%          & 59.59\%   & +12.02\% \\
\hline + $[A+B]$ and $[|A-B|]$            & 62.71\% & 70.37\% & 61.04\% & 69.02\%          & 65.79\%   & +6.20\%\\
\hline + $[A \cdot B]$                    & 66.20\% & 68.61\% & 60.40\% & 70.18\%          & 66.35\%   & +0.56\% \\
\hline + $[m_2\cdot [A, B]]$ & 65.73\% & 70.74\% & 60.90\% & \textbf{70.97\%} & 67.09\%   & +0.74\% \\
\hline
\end{tabular}
\end{table*}

Our method was very scalable. It took on average several hours to
generate the word embedding file from our medical corpus with 20G
text using $16\times3.2$G cpus and roughly 30 minutes to finish
the training process using one cpu. To measure the scalability at
the apply time, we constructed a new medical synonym knowledgebase
with our best synonym extractor. This was done by applying the
concept space model trained under the NEG+SKIP setting to a set of
11B pairs of terms. All these terms are associated with CUIs, and
occur at least twice in our medical corpus. This KB construction
process finished in less than 10 hours using one cpu, resulting in
more than 3M medical synonym term pairs. To evaluate the recall of
this knowledgebase, we checked each term pair in the held out
synonym dataset against this KB, and found that more than $42\%$
of them were covered by this new KB. Precision evaluation of this
KB requires a lot of manual annotation effort, and will be
included in our future work.

\subsection{Feature Contribution Analysis}\label{sec:contributionresults}
In Section~\ref{sec:featureexpansion}, we expand the raw features
with the matching features and several other feature expansions to
model the term relationships. In this section, we study the
contribution of each individual feature to the final results. We
added all those expanded features to the raw features one by one
and re-ran the experiments for the concept space model. The
results and the feature contributions are summarized in
Table~\ref{tbl:featurecontributions}.

The results show that adding matching features, ``sum'' and
``difference'' features can significantly improve the $F_1$
scores. We can also see that adding the last two feature sets does
not seem to contribute a lot to the average $F_1$ score. However,
they do contribute significantly to our best $F_1$ score by about
$2\%$ .

\section{Conclusions}
In this paper, we present an approach to construct a medical
concept space from manually extracted medical knowledge and a
large corpus with 20G unstructured text. Our approach extends the
Word2Vec model by making use of the medical knowledge as extra
label information during the training process. This new approach
fits well for the medical domain, where the language use
variability is exceptionally high and the existing knowledge is
also abundant.

Experiment results show that the proposed model outperforms the
baseline approaches by a large margin on a dataset with more than
one million term pairs. Future work includes doing a precision
analysis of the resulting synonym knowledgebase, and exploring how
deep learning models can be combined with our concept space model
for better synonym extraction.

\end{document}